\documentclass{article}

\usepackage{arxiv}

\usepackage[utf8]{inputenc} 
\usepackage[T1]{fontenc}    
\usepackage{hyperref}       
\usepackage{url}            
\usepackage{booktabs}       
\usepackage{amsfonts}       
\usepackage{nicefrac}       
\usepackage{microtype}      
\usepackage{lipsum}
\usepackage{graphicx}

\usepackage{algorithm}
\usepackage{algorithmic}
\usepackage{amsmath}
\usepackage{amssymb}
\usepackage{multirow}
\usepackage{tabularx}
\usepackage{boldline}

\graphicspath{ {./images/} }

\title{Self-Supervised Contrastive Learning for Videos using Differentiable Local Alignment}

\author{
 Keyne Oei \\
  Universität des Saarlandes\\
  Saarbrücken, Germany \\
  \texttt{s8keoeii@uni-saarland.de} \\
   \And
 Amr Gomaa \\
  German Research Center for Artificial Intelligence (DFKI)\\
  Saarbrücken, Germany\\
  \texttt{amr.gomaa@dfki.de} \\
  \And
 Anna Maria Feit \\
  Universität des Saarlandes\\
  Saarbrücken, Germany \\
  \texttt{feit@cs.uni-saarland.de} \\
  \And
 João Belo \\
  Universität des Saarlandes\\
  Saarbrücken, Germany \\
  \texttt{jbelo@cs.uni-saarland.de} \\
}

\begin{document}
\maketitle
\begin{abstract}
    Robust frame-wise embeddings are essential to perform video analysis and understanding tasks. We present a self-supervised method for representation learning based on aligning temporal video sequences. Our framework uses a transformer-based encoder to extract frame-level features and leverages them to find the optimal alignment path between video sequences. We introduce the novel Local-Alignment Contrastive (LAC) loss, which combines a differentiable local alignment loss to capture local temporal dependencies with a contrastive loss to enhance discriminative learning. Prior works on video alignment have focused on using global temporal ordering across sequence pairs, whereas our loss encourages  identifying the best-scoring subsequence alignment. LAC uses the differentiable Smith-Waterman (SW) affine method, which features a flexible parameterization learned through the training phase, enabling the model to adjust the temporal gap penalty length dynamically. Evaluations show that our learned representations outperform existing state-of-the-art approaches on action recognition tasks.
\end{abstract}


\section{Introduction}

Video understanding and analysis have garnered significant attention in computer vision research in recent years \cite{2016_tsn, 2017_quovadis, 2018_trn, 2019_slowfast, 2019_actiontransnet, 2020_x3d, 2021_vit, 2021_vivit, 2021_videoclass_actionclip, 2024_astar}. 
These tasks require models to capture not only spatial information within individual frames but also the complex temporal dynamics that evolve across sequences of frames. 
Traditional supervised learning methods have achieved considerable success in advancing performance; 
models such as those presented in \cite{2017_quovadis, 2019_slowfast, 2024_astar} are able to predict action categories effectively.
However, these methods present several challenges: (i) Most approaches require fine-grained annotations, which are time-consuming and labor-intensive; (ii) Most approaches struggle to capture the complex temporal information and causal relationships that vary across diverse videos.
To address these challenges, we explore how representation learning based on the pretext-task of video alignment can improve performance in downstream action recognition tasks such as phase classification and phase progression.

Prior self-supervised or weakly-supervised video alignment approaches \cite{2017_tcn, 2019_tcc, 2021_lav, 2021_gta, 2022_carl, 2024_lrprop} train on pairs of videos that describe the same action using either cycle-consistency \cite{2019_tcc}, soft dynamic time warping \cite{2021_gta, 2021_lav}, or sequential contrastive learning \cite{2022_carl}.
We notice that the pretext task of video alignment shares similarities with bioinformatics sequence alignment for studying DNA, RNA, and protein structures. 
Since some actions are composed of sub-actions or events that occur in a specific order, video alignment shares similarities with protein alignment, as both involve identifying regions of continuity and discontinuity within sequences.
There are two main types of alignment algorithms commonly used in bioinformatics: \textit{global} and \textit{local}. 
The Needleman-Wunsch (NW) algorithm \cite{1970_nw}, a global alignment method, aligns entire sequences from start to finish. 
This approach is conceptually similar to Dynamic Time Warping (DTW) \cite{2007_dtw}, which aligns sequences of varying lengths by matching their temporal patterns from start to finish.
Differentiable versions of NW and DTW algorithms have been established by incorporating smooth maximum/minimum and argmax/argmin functions \cite{2020_diff_nw, 2017_softdtw}.
In particular, Soft-DTW \cite{2017_softdtw} has been used in prior video alignment approaches \cite{2021_lav, 2021_gta, 2024_lrprop}.
In contrast, the Smith-Waterman (SW) algorithm~\cite{1981_sw} implements local alignment, identifying regions of high similarity within sequences and allowing for gaps, insertions, and deletions.
This ability to handle subsequences makes local alignment particularly suitable for modeling the complexity of actions in real-world video data, where actions may not always follow rigid, continuous sequences.

\begin{figure}[t]
\begin{tabular}{c}
\includegraphics[width=\textwidth]{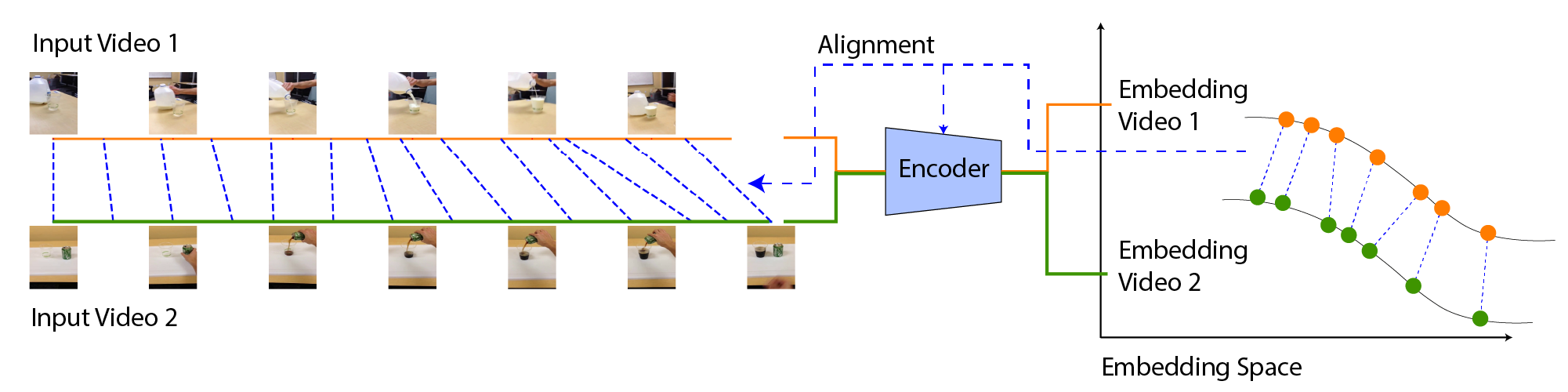}
\end{tabular}
\caption{We introduce a representation learning approach that aligns video sequences depicting the same processes. Our training objective is to use a novel LAC loss to optimize and learn an element-wise embedding function that supports the alignment process.}
\label{fig: vis_intro}
\end{figure}

Specifically, in this work, we introduce a novel loss function termed \textit{Local-Alignment Contrastive} (LAC) loss, integrated into our proposed end-to-end framework for the video alignment task, as visualized in Figure \ref{fig: vis_intro}.
Following prior works \cite{2022_carl, 2024_lrprop}, our framework uses a variation of convolutional and transformer encoders to extract spatio-temporal features from each frame. 
However, rather than relying on a differentiable global alignment loss like Soft-DTW \cite{2017_softdtw}, our novel LAC loss enables the comparison of video pairs through a consistent and differentiable local alignment loss. 
This loss specifically accommodates sequences with learned penalties for opening gaps and extending them, combined with a contrastive loss that effectively separates dissimilar frames.
We demonstrate that the learned representations of our approach outperform previous methods on various downstream action recognition tasks, such as action phase classification and action phase progression on the Pouring~\cite{2017_tcn} and PennAction~\cite{2013_pennaction} datasets. 
Code and models are available at: \href{https://github.com/keynekassapa13/LAC}{https://github.com/keynekassapa13/LAC}.

\section{Related Work}

\textbf{Action Recognition.} Over the last decade, researchers have proposed various methods to address video recognition tasks~\cite{2017_quovadis, 2019_slowfast, 2016_tsn, 2018_trn, 2019_actiontransnet}, including action classification~\cite{2014_vc_karpathy, 2014_vc_simonyan, 2017_quovadis, 2020_x3d, 2021_vivit, 2021_videoclass_actionclip}, which involves mapping a video to an action category, and action segmentation~\cite{2017_actiondetect, 2018_videocapsulenet, 2019_slowfast, 2018_bsn, 2022_actionformer}, which aims to identify the spatial and temporal boundaries of specific actions within a video.
However, these methods often rely heavily on extensive labeled data, which is time-consuming and labor-intensive to obtain. 
To address this limitation, self-supervised learning methods have been explored, enabling the extraction of video representations without requiring labeled data. 
Video alignment, as a self-supervised pretext task for action recognition, leverages temporal consistency to learn meaningful representations. 
Previous work has utilized traditional computer vision techniques to derive these representations; for instance, optical flow~\cite{1994_opticalflow} computes motion between frames based on brightness constancy, the CONDENSATION algorithm~\cite{1998_condensation} employs a probabilistic approach informed by prior motion data, and Space-Time Interest Points (STIPs)~\cite{2005_stips} detect salient points by analyzing both spatial and temporal patterns in video sequences.

\noindent \textbf{Self-Supervised Learning.} With the rise of deep learning, the field has shifted towards using Self-Supervised Learning (SSL) methods to learn video representations. Prior works \cite{2017_tcn, 2019_tcc, 2021_lav, 2021_gta, 2022_carl} have addressed the problem of video alignment using self-supervised methods. Time Contrastive Network (TCN) \cite{2017_tcn} utilizes contrastive learning to distinguish frames from different segments while grouping those within the same segment. Temporal Cycle-Consistency (TCC) \cite{2019_tcc} introduces cycle-consistency loss to identify and match recurring action sequences within or across videos. Learning by Aligning Videos (LAV) \cite{2021_lav} uses Soft-DTW \cite{2017_softdtw} and Inverse Difference Moment (IDM) regularization to optimize alignment and ensure balanced frame distribution. Global Temporal Alignment (GTA) \cite{2021_gta} implements a modified differentiable DTW with global consistency loss for consistent temporal alignment. Contrastive Action Representation Learning (CARL) \cite{2022_carl} introduces Sequence Contrastive Loss (SCL), which minimizes the KL-divergence between augmented views based on timestamp distance. Learning Representation by position PROPagation (LRProp) \cite{2024_lrprop} utilizes a Soft-DTW \cite{2017_softdtw} alignment path and a novel pair-wise position propagation technique. We will use these state-of-the-art methods to evaluate our model's performance on action recognition tasks.

\noindent \textbf{Global vs. Local Alignment.} 
Global alignment methods, such as the NW algorithm \cite{1970_nw} and DTW \cite{2007_dtw}, aim to align entire sequences from start to finish to achieve the best overall match; differentiable versions have been developed \cite{2017_softdtw, 2020_diff_nw} and utilized in prior video alignment approaches \cite{2021_lav, 2021_gta, 2024_lrprop}. 
In contrast, the SW algorithm \cite{1981_sw} focuses on identifying the most similar subsequences within larger sequences.
Recent work by Petti et al. \cite{2023_sw1} introduced a differentiable SW algorithm in a supervised framework that jointly learns alignments and improves protein structure prediction.
\cite{2023_sw2} shows the power of embedding sequences into high-dimensional spaces within a supervised framework for more precise local alignment.
Our approach employs a differentiable version of SW in self-supervised framework to improve model performance on video understanding and alignment tasks compared to state-of-the-art temporal global alignment methods.
\section{Method}

\begin{figure}[t]
\includegraphics[width=1\textwidth]{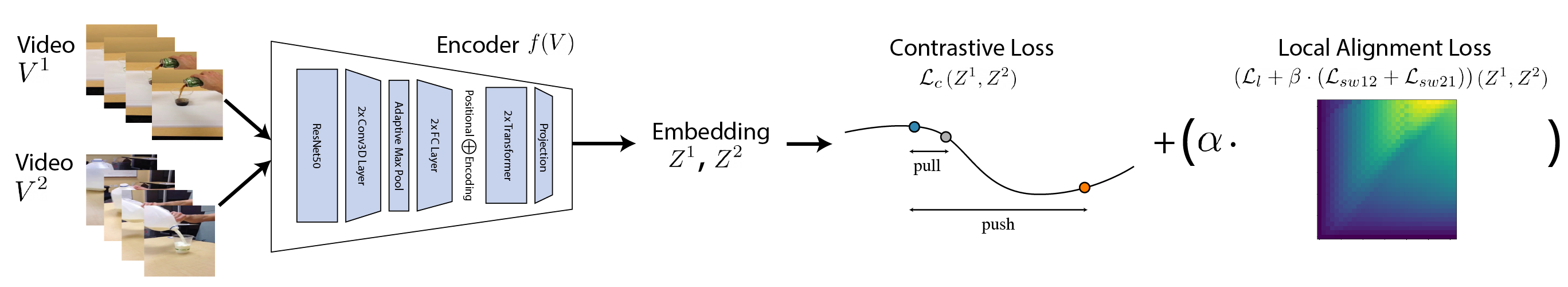}
\caption{Our framework uses Local Alignment Loss and Contrastive Loss to optimize embeddings generated by an encoder that processes input videos after they have undergone spatio-temporal data augmentation.}
\label{fig: vis_method}
\end{figure}

This section introduces our novel Local-Alignment Contrastive (LAC) loss function employed within our framework where video alignment serves as the pretext task, as visualized in Fig. \ref{fig: vis_method}. Specifically, we derive the forward and backward passes of the differentiable local alignment loss and explain how we ensure consistency within our method.

\subsection{Notations} \label{sec: 3_notation}

Let $V^1$ and $V^2$ represent two videos, each consisting of a sequence of frames. 
Specifically, $V^1 = \{ v^1_t \}_{t=1}^{T_1}$ and $V^2 = \{ v^2_t \}_{t=1}^{T_2}$, where $v^1_t$ and $v^2_t$ denote individual frames at time $t$, and $T_1$ and $T_2$ are the total number of frames in each video. 
Following \cite{2022_carl}, we preprocess the videos using temporal random cropping to generate cropped versions $V_c^1$ and $V_c^2$. 
The indices of the sampled frames during cropping are recorded in $S^1$ and $S^2$, where $S^1 \subseteq \{1, \dotsc, T_1\}$ and $S^2 \subseteq \{1, \dotsc, T_2\}$. 
Next, we apply data augmentations to the cropped videos, resulting in \( \widetilde{V}^1 \) and \( \widetilde{V}^2 \). 
These augmented videos are then processed through an encoder to obtain embeddings \( Z^1 \) and \( Z^2 \). 
Each embedding \( Z^i \in \mathbb{R}^{T_i \times E} \) represents a sequence of length \( T_i \) with an embedding dimension \( E \). 
The objective is to train the encoder to minimize the distance between embeddings \( Z^1 \) and \( Z^2 \) corresponding to videos depicting the same action, as determined by our loss.

\subsection{Background}

Prior approaches \cite{2021_lav, 2021_gta} use DTW as the main method for the temporal alignment loss. 
Given embeddings \( Z^1 \) and \( Z^2 \), we compute a cost matrix \( C \in \mathbb{R}^{T_1 \times T_2} \), where each element \( C_{i,j} \) measures the local distance between \( Z^1_i \) and \( Z^2_j \).
DTW seeks the optimal alignment path through this cost matrix that minimizes the total cumulative cost.
The feasible paths must satisfy constraints of matching endpoints, monotonicity, and continuity. 
While the number of possible alignment paths grows exponentially with the lengths of the sequences, dynamic programming allows DTW to compute the optimal path efficiently in quadratic time \( O(T_1 T_2) \). 
The DTW algorithm is formulated recursively as:

\begin{equation}
    D_{i, j} = C_{i, j} + \min \left\{
    \begin{array}{l}
         D_{i-1, j-1} \\
         D_{i-1, j} \\
         D_{i, j-1} \\
    \end{array}
    \right.
    \label{eq: dtw}
\end{equation}

where \( D_{i, j} \) represents the accumulated cost up to point \( (i, j) \), and \( C_{i, j} \) is the local cost between elements \( Z^1_i \) and \( Z^2_j \).
The non-differentiable nature of the DTW formulation presents challenges for gradient-based optimization techniques. 
To address this, prior methods \cite{2021_lav, 2021_gta} utilize Soft-DTW \cite{2017_softdtw} which replaces the discrete \( \min \) operator with a differentiable, smoothed version denoted as \( \text{softmin}^\gamma \).
As the smoothing parameter \( \gamma \) approaches zero, the behavior of \( \text{softmin}^\gamma \) approximates that of the discrete \( \min \) operator. 
This substitution enables the use of gradient-based optimization in DTW while preserving its alignment capabilities.

\subsection{Differentiable Local-Alignment} \label{sec: 3_diffla}

Despite the advantages offered by Soft-DTW \cite{2017_softdtw}, it inherently maintains the global alignment characteristic, ensuring continuity throughout the entire sequence. This attribute of global alignment may prove suboptimal in scenarios where localized alignment strategies, such as those implemented by the SW algorithm \cite{1981_sw}, are preferable. The SW algorithm \cite{1981_sw} supports localized matching, providing enhanced flexibility and increased sensitivity to similarities within subsequences. Therefore, we propose a differentiable SW local-alignment approach as our method for temporal alignment loss.

We consider a pair of embeddings, $(Z^1, S^1)$ and $(Z^2, S^2)$, with corresponding sampled indices, tasked with determining the optimal alignment between these embeddings.
We begin by calculating a similarity matrix, $S \in \mathbb{R}^{T \times T}$, using the inverted Euclidean distance to yield higher values that indicate increased similarity between the points in $Z^1$ and $Z^2$. 
Utilizing dynamic programming, the SW algorithm effectively determines the optimal alignment by maximizing a score derived from pairwise similarities within the matrix. 
The alignment algorithm is formally expressed as follows:

{
\begin{equation}
    D_{i,j} = S_{i, j} + \max \left\{
    \begin{array}{l}
        0 \\
        D_{i - 1,j - 1} \\
        I_{x_{i - 1,j - 1}} \\
        I_{y_{i - 1,j - 1}}
    \end{array} \label{eq: la_D}
\right. \\
\end{equation}

\begin{equation}
I_{x_{i,j}} = \max \left\{
    \begin{array}{l}
        D_{i, j - 1} - g_{o_{i, j}} \\
        I_{x_{i, j -1}} - g_{e_{i, j}}
    \end{array}
\right. \label{eq: la_Ix}
\end{equation}

\begin{equation}
I_{y_{i,j}} = \max \left\{
    \begin{array}{l}
        D_{i - 1, j} - g_{o_{i, j}}  \\
        I_{x_{i - 1, j}} - g_{o_{i, j}} \\
        I_{y_{i - 1, j}} - g_{e_{i, j}} 
    \end{array}
\right. \label{eq: la_Iy}
\end{equation}
}

where the initial values are set to $D(i, 0) = D(0, j) = I_x(i, 0) = I_x(0, j) = I_y(i, 0) = I_y(0, j) = -\infty$ for all indices $i, j = 1, \ldots, T$. The matrix $D$ described in Eq. \ref{eq: la_D} acts as the primary scoring matrix, representing the cumulative maximum score at each matrix coordinate $(i, j)$. The matrices $I_x$ and $I_y$ detailed in Eq. \ref{eq: la_Ix} and Eq. \ref{eq: la_Iy}, represent the scores associated with introducing gaps along the $x$ and $y$ axes. Additionally, $g_o$ and $g_e$ are used to specify the penalties for opening and extending gaps.

Similar to Soft-DTW \cite{2017_softdtw}, our differentiable temporal alignment loss function uses a smoothed version of the discrete $\max$ function, denoted as $\max^\gamma$. This smooth approximation is formally defined as:

\begin{equation}
  \max{}^{\gamma}(u_1, \ldots, u_n) := \gamma \, \log \, \sum_{i=1}^{n} \, e^{u_i/\gamma}  \label{eq: smooth_max}
\end{equation}

where $\gamma > 0$ is a smoothing parameter that controls the approximation's fidelity to the original $\max$ function. As $\gamma$ approaches zero, the smoothed $\max^\gamma$ function increasingly approximates the behavior of the standard $\max$ function, enabling a differentiable formulation that can be integrated into gradient-based optimization frameworks. The gradient of the smoothed $\max^\gamma$ with respect to each input $u_i$ is calculated as follows:

\begin{align}
    \frac{\partial \max{}^\gamma(\{u_1,\ldots, u_n\})}{\partial u_i} & = \exp\left(\frac{u_i - \max{}^\gamma(\{u_1,\ldots,u_n\})}{\gamma}\right)
    \label{eq: partial_softmax}
\end{align}

This expression aligns with the softmax function. 
By differentiating the smoothed maximum score with respect to each input, our model directly utilizes derivative-sensitive parameters such as the similarity score ($S$), gap open penalty ($g^o$), and gap extend penalty ($g^e$). 
As a result, we define our differentiable local alignment loss as \(\mathcal{L}_{sw_{ij}} = \text{SW}(\text{sim}(Z^i, Z^j))\).
The ($\text{sim}$) function calculates the similarity between two embeddings and generates a similarity matrix $S$. 
Unlike prior differentiable SW implementations \cite{2023_sw1, 2023_sw2} that depend on supervised learning, our approach integrates the differentiable SW algorithm within a self-supervised framework.

\subsection{Final Loss} \label{sec: 3_total}

Following the approach in \cite{2022_carl}, our contrastive loss is modeled after the NT-Xent loss from SimCLR \cite{2020_simclr}. 
This loss function calculates absolute pairwise distances between embeddings $(Z^1, S^1)$ and $(Z^2, S^2)$,  subsequently forming a Gaussian-weighted positive label distribution.
This distribution is then contrasted against normalized similarity logits using the Kullback-Leibler divergence. 
The contrastive loss is mathematically expressed as:

\begin{equation}
    \mathcal{L}_c = - \frac{1}{T} \sum_{i, j=1}^T \left( \frac{G(s_i^1 - s_j^2)}{\sum_{k=1}^T G(s_i^1 - s_k^2)} \log \frac{\exp(\text{sim}(z_i^1, z_j^2) / \tau)}{\sum_{k=1}^T \exp(\text{sim}(z_i^1, z_k^2) / \tau)} \right)
\end{equation}

where $\tau$ represents the temperature parameter, $G$ denotes the Gaussian weighting function applied to the absolute pairwise distances, and $(\text{sim})$ measures the normalized similarity between embeddings.
To ensure consistency between contrastive and local alignment losses, we synchronize our temporal local loss by utilizing the primary score matrix $D_{12}$ and $D_{21}$ from $\mathcal{L}_{sw12}$ and $\mathcal{L}_{sw21}$.
Following the application of the softmax function to $D$, the resulting matrix is referred to as $\Tilde{D}$.
The logits matrix \(L\) is then derived by performing an element-wise multiplication of \(\Tilde{D}_{12}\) and \(\Tilde{D}_{21}^T\). 
The cross-entropy loss for this logits matrix against the Gaussian-weighted labels is calculated as follows:

\begin{align*}
\mathcal{L}_l &= -\frac{1}{T} \sum_{i, j=1}^T  \left(\, \frac{G(s_i^1 - s_j^2)}{\sum_{k=1}^T G(s_i^1 - s_k^2)} \log \left( \frac{\exp(\text{L}_{ij})}{\sum_{k=1}^T \exp^(\text{L}_{ik})} \right) \right), \\
\text{where} 
\,\ \Tilde{D}_{12, ij} & = \frac{\exp(D_{12, ij}/\tau)}{\sum_{k=1}^T \exp(D_{12, ik}/\tau)} \,,
\,\ \Tilde{D}_{21, ij} = \frac{\exp(D_{21, ij}/\tau)}{\sum_{k=1}^T \exp(D_{21, ik}/\tau)} \,,
\,\ L = \Tilde{D}_{12} \cdot \Tilde{D}_{21}^T
\end{align*}

Our proposed loss function, referred as Local-Alignment Contrastive (LAC), integrates contrastive and local alignment losses. It is formally defined as follows:

\begin{equation}
    \mathcal{L} = \mathcal{L}_c + \alpha \cdot (\mathcal{L}_l + \beta \cdot (\mathcal{L}_{sw12} + \mathcal{L}_{sw21}))
\end{equation}

where \(\alpha\) and \(\beta\) are weights used to balance the components of the loss, set to 0.01 and 1, respectively.


\section{Experiments}

\textbf{Datasets.} We evaluate the performance of the LAC model on two datasets using a variety of evaluation metrics. The Pouring dataset \cite{2017_tcn}, which focuses on the action of pouring liquids, includes 70 training and 14 testing videos. The PennAction dataset \cite{2013_pennaction}, featuring 13 human actions, contains 1140 training and 966 testing videos. We utilize the key events and phases for the videos in both datasets as proposed by TCC \cite{2019_tcc}.

\noindent \textbf{Implementation Details.} 
Our encoder, \(f: \mathbb{R}^{TxCxWxH} \to Z\), maps video inputs $V$ into an embedding space \(Z\). We use a ResNet50-v2 \cite{2015_resnet} as our backbone to extract features from the Conv4c layer with an output size of \(10x10x512\).
These features are then processed through adaptive max pooling, followed by two fully connected layers with ReLU activation.
A subsequent linear layer projects the features into a 256-dimensional space. 
To enhance the model’s capacity to capture long-range dependencies, we integrate sine-cosine positional encoding and employ a two-layer Transformer encoder. 
To improve the model’s ability to capture long-range dependencies, we incorporate sine-cosine positional encoding and apply a two-layer Transformer encoder. 
The final embedding layer reduces the dimensionality to 128 for the frame-wise representations.

\begin{table}[]
\small
\centering
\begin{tabularx}{\textwidth}{X|ccc|ccc|c|c}
\multirow{2}{*}{\textbf{Method}} & \multicolumn{3}{c|}{\textbf{Class}} & \multicolumn{3}{c|}{\textbf{AP@K}} & \multirow{2}{*}{\textbf{Progress}} & \multirow{2}{*}{\textbf{$\tau$}} \\ 
\cline{2-7}
 & \textbf{10} & \textbf{50} & \textbf{100} & \textbf{K=5} & \textbf{K=10} & \textbf{K=15} & & \\ 
\hline
\hline
TCN \cite{2017_tcn}        & 80.32 & 81.44 & 83.56 & 76.26 & 76.71 & 77.26 &  82.30 & 83.51 \\
TCC \cite{2019_tcc}        & 86.60 & 86.78 & 86.86 & 81.84 & 80.94 & 81.69 &  83.36 & 85.26 \\
LAV \cite{2021_lav}        & 89.77 & 90.35 & 91.77 & 87.48 & 88.36 & 88.40 &  85.20 & 88.75 \\
GTA \cite{2021_gta}        & 89.34 & 90.20 & 90.22 & 87.79 & 87.48 & 87.82 &  88.67 & 92.47 \\
SCL \cite{2022_carl}       & \underline{92.76} & 92.80 & 93.05 & 88.75 & 88.51 & 88.97 & 91.26 & \underline{98.20} \\
LRPROP \cite{2024_lrprop}  & 92.70 & \underline{94.44} & \underline{94.36} & \underline{92.41} & \underline{90.33} & \underline{90.86} & \underline{94.09} & \textbf{99.46} \\
\hline
LAC & \textbf{95.87} & \textbf{95.78} & \textbf{95.16} & \textbf{92.76} & \textbf{91.07} & \textbf{91.37} & \textbf{94.24} & 97.50 \\
\hline
\hline
\end{tabularx}
\caption{Performance comparison of state-of-the-art methods on the Pouring Dataset \cite{2017_tcn}}
\label{tab: p_class}
\end{table}

\noindent \textbf{Evaluation Metrics.}
Following related work \cite{2019_tcc, 2021_lav, 2021_gta, 2022_carl, 2024_lrprop}, we evaluate our model using the following metrics:
(i) \textit{Phase Classification}, which assesses the accuracy of action phase predictions by training an SVM classifier on our embeddings;
(ii) \textit{Phase Progression}, which evaluates how accurately our embeddings predict action progress using a linear regression model's average R-squared value, based on normalized timestamp differences;
(iii) \textit{Average Precision@K (AP@K)}, which evaluates fine-grained frame retrieval accuracy by calculating the proportion of correctly matched phase labels within the K closest frames;
(iv) \textit{Kendall's Tau}, which quantifies the temporal alignment between sequences by comparing the ratio of concordant to discordant frame pairs.

\subsection{Results}

We apply the same four metrics to the Pouring dataset \cite{2017_tcn}. For the PennAction dataset \cite{2013_pennaction}, we evaluate each of the 13 action categories using the same four metrics and report the averaged results across all categories.
To ensure a fair comparison (e.g., consistent data splits, identical preprocessing), we replicated the evaluations of previous approaches using the GitHub repositories \footnote{github.com/google-research/google-research/tcc, github.com/trquhuytin/LAV-CVPR21,\\github.com/hadjisma/VideoAlignment, github.com/minghchen/CARL\_code} provided by the original authors. 
Each model was trained using its respective pre-trained backbone. 
An exception was made for LRPROP, as their GitHub repository is not available.

\noindent \textbf{Results on Pouring Dataset.} 
Table~\ref{tab: p_class} presents a comparison of our method's performance against state-of-the-art approaches on the Pouring dataset. Bold and underlined text denote the best and second-best results.
Notably, it achieves a +3.11\% improvement on Phase Classification using only 10\% of the labels. Additionally, our model excels in AP@K and Progress metrics.
However, we observe lower performance in Kendall's Tau. We hypothesize this is due to how the SW's algorithm encourages skipping unnecessary segments of the sequence. The introduction of gap open and extend penalties could disrupt the continuity needed for high Kendall's Tau scores.

\begin{table}[]
\centering
\begin{tabularx}{\textwidth}{X|ccc|ccc|c|c}
\multirow{2}{*}{\textbf{Method}} & \multicolumn{3}{c|}{\textbf{Class}} & \multicolumn{3}{c|}{\textbf{AP@K}} & \multirow{2}{*}{\textbf{Progress}} & \multirow{2}{*}{\textbf{$\tau$}} \\ 
\cline{2-7}
 & \textbf{10} & \textbf{50} & \textbf{100} & \textbf{K=5} & \textbf{K=10} & \textbf{K=15} & & \\ 
\hline
\hline
TCN \cite{2017_tcn}        & 69.73 & 70.26 & 70.01 & 60.92 & 61.57 & 61.52 &  76.37 & 63.72 \\
TCC \cite{2019_tcc}        & 86.60 & 86.78 & 86.86 & 81.84 & 80.94 & 81.69 &  83.36 & 85.26 \\
LAV \cite{2021_lav}        & 88.51 & 88.72 & 88.97 & 73.47 & 73.13 & 74.27 &  92.52 & 93.06 \\
GTA \cite{2021_gta}        & 84.21 & 84.68 & 85.28 & 71.72 & 72.17 & 71.52 &  90.51 & 83.35 \\
SCL \cite{2022_carl}       & 87.85 & 87.52 & 88.15 & 91.70 & 90.61 & 90.58 &  92.89 & \underline{98.14}\\
LRPROP \cite{2024_lrprop}  & \underline{91.90} & \underline{92.96} & \underline{93.25} & \underline{92.46} & \underline{92.2} & \underline{92.03} & \underline{93.03} & \textbf{99.09} \\
\hline
LAC & \textbf{95.57} & \textbf{93.79} & \textbf{93.40} & \textbf{93.87} & \textbf{93.41} & \textbf{92.65} & \textbf{94.21} & 94.10 \\
\hline
\hline
\end{tabularx}
\caption{Performance comparison of state-of-the-art methods on the PennAction Dataset \cite{2013_pennaction}}
\label{tab: pn_class}
\end{table}

\begin{figure}[t]
\begin{tabular}{cc}
\includegraphics[width=0.195\textwidth]{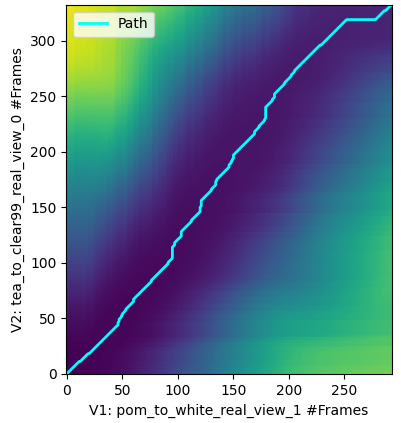}&\includegraphics[width=0.764\textwidth]{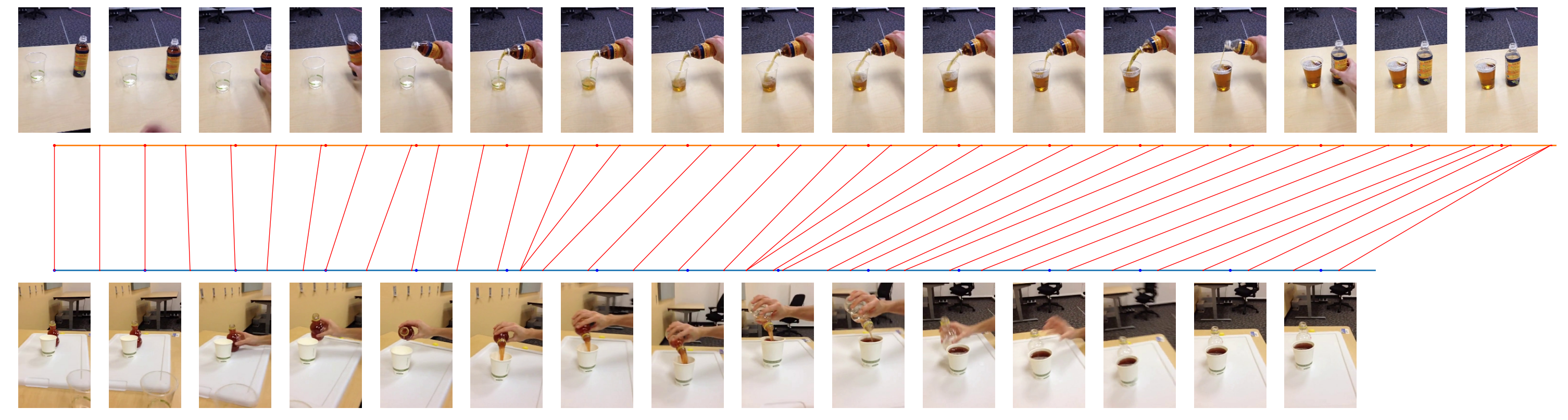}
\end{tabular}
\caption{Similarity matrix (left) shows video alignment using an optimal path and the respective video aligned frame-by-frame (right).}
\label{fig: vis_align}
\end{figure}

\noindent \textbf{Results on PennAction Dataset.} 
Table \ref{tab: pn_class} compares the performance of our method with state-of-the-art approaches on the PennAction dataset. Bold and underlined text denote the best and second-best results. 
LAC consistently outperforms previous methods across most metrics, with the exception of Kendall's Tau. 
Notably, the improvement in AP@K is more pronounced on PennAction than on Pouring, likely due to the fewer number of frames in PennAction dataset, which may further impact alignment performance.

\begin{figure}[t]
\begin{tabular}{cc}
\includegraphics[width=0.45\textwidth]{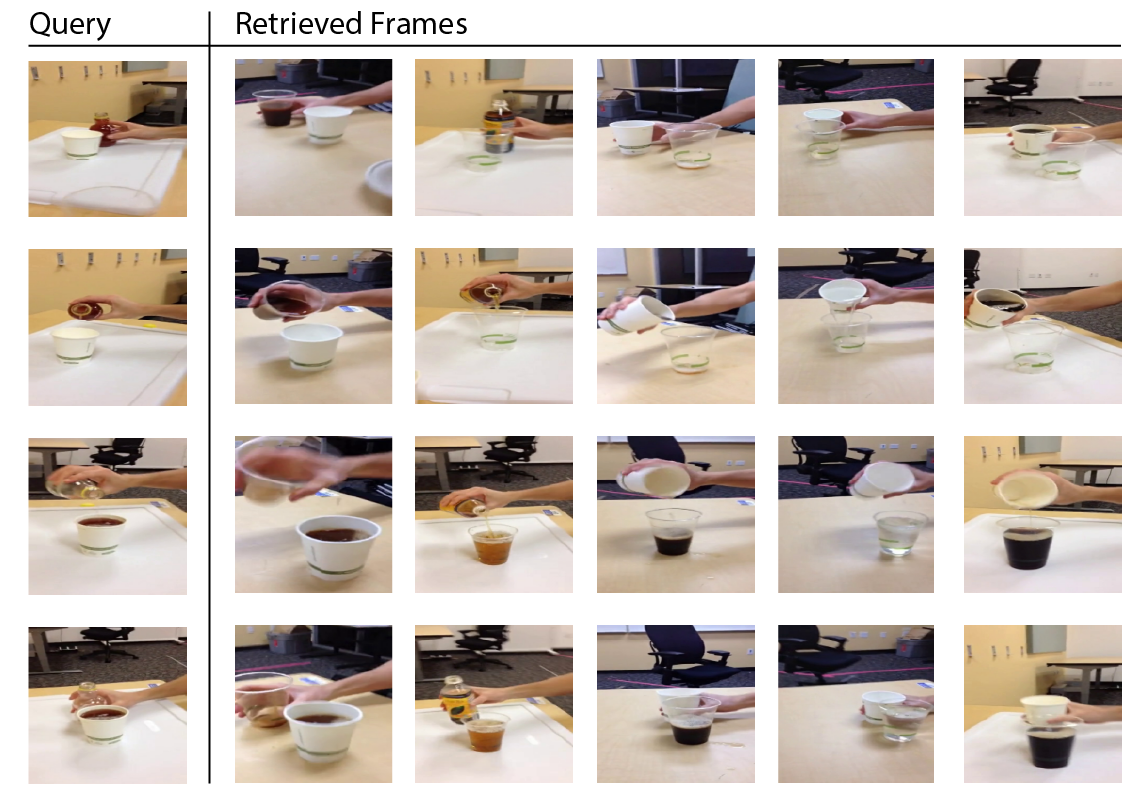}&\includegraphics[width=0.45\textwidth]{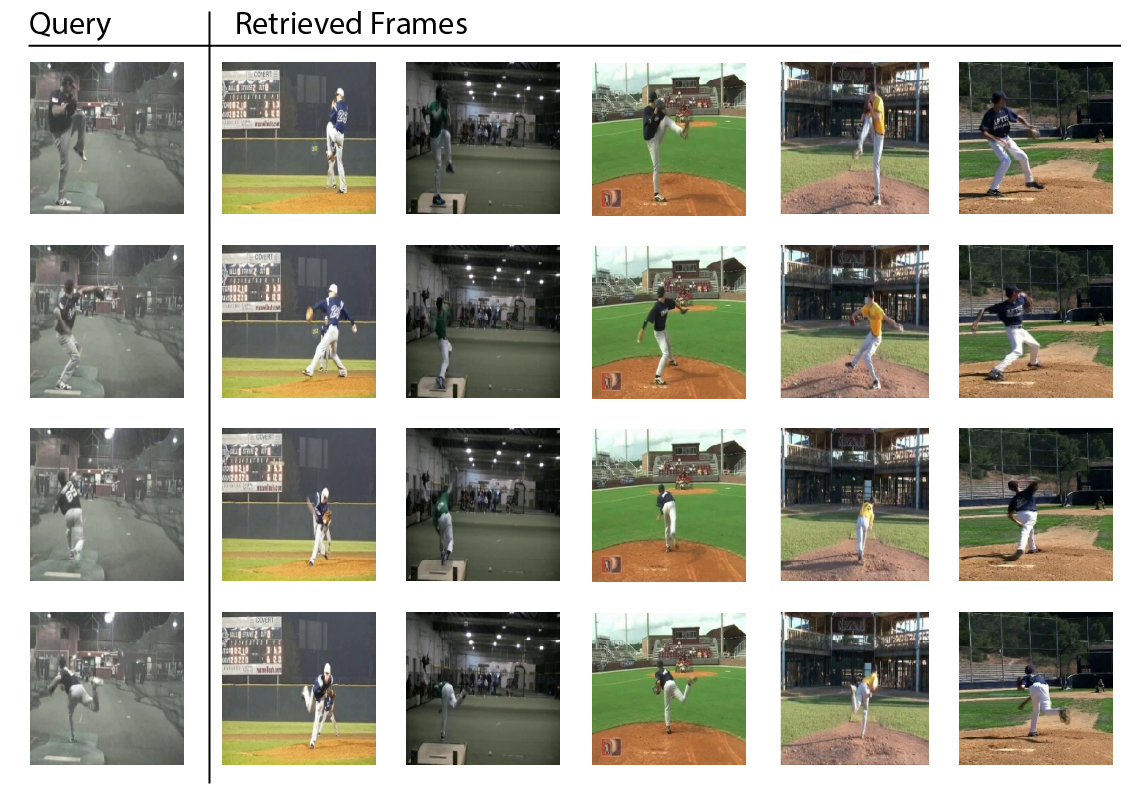}
\end{tabular}
\caption{Fine-grained frame retrieval for Pouring (left) and PennAction (right) achieved using nearest neighbors within our embedding space.}
\label{fig: vis_retr}
\end{figure}

\noindent \textbf{Qualitative analysis of results.} 
Figure \ref{fig: vis_align} illustrates the alignment process, with the optimal path depicted on the left and the frame-by-frame alignment on the right. 
This visualization demonstrates the synchronization of the two videos despite differences in their temporal progression and duration. 
Such alignment visualizations enable in-depth analysis of deviations or inefficiencies within specific actions in video understanding.
Additionally, Figure \ref{fig: vis_retr} showcases our embedding-based retrieval system's ability to accurately identify and retrieve frames corresponding to specific action sequences across different videos.

\subsection{Ablation Study}

This section presents multiple experiments on the \textit{Pouring} dataset that analyze the different components of our framework. 

\begin{table}[H]
\centering
\begin{tabularx}{0.99\linewidth}{|X|ccc|}
\hline
\textbf{Architecture} & \textbf{Class} & \textbf{Progress} & \textbf{$\tau$} \\
\hlineB{3}
ResNet-50 + Convolutional 3D & 88.04 & 73.26 & 71.37 \\
\hline
ResNet-50 + Tranformer 
(w/o pretrained weights) & 90.06 & 89.32 & 94.87 \\
\hline
ResNet-50 + Transformer
 & \textbf{95.16} & \textbf{94.24} & \textbf{97.50} \\
\hline
\end{tabularx}
\caption{Different Encoder Architecture on Model Performance.}
\label{tab_lac_arch}
\end{table}

\begin{table}[H]
\parbox{.4\linewidth}{
\centering
\begin{tabularx}{0.99\linewidth}{|X|ccc|}
\hline
\textbf{$\gamma$} & \textbf{Class} & \textbf{Progress} & \textbf{$\tau$} \\
\hlineB{3}
0.6 & 93.44 & 92.12 & 94.93 \\
\hline
0.7 & 94.87 & 92.82 & 97.29 \\
\hline
0.8 &  \textbf{95.16} & \textbf{94.24} & \textbf{97.50} \\
\hline
0.9 & 93.93 & 92.02 & 94.91 \\
\hline
\end{tabularx}
\caption{Different $\gamma$ Values on Model Performance}
\label{tab_lac_y}
}
\hfill
\parbox{.59\linewidth}{
\centering
\begin{tabularx}{0.99\linewidth}{|X|ccc|}
\hline
\textbf{Loss $\mathcal{L}$} & \textbf{Class} & \textbf{Progress} & \textbf{$\tau$} \\
\hlineB{3}
$\mathcal{L}_c $ & 93.44 & 92.12 & 94.93 \\
\hline
$\mathcal{L}_c + \alpha \cdot \mathcal{L}_l$ & 93.71 & 92.16 & 95.01 \\
\hline
$\mathcal{L}_c + \alpha \cdot (\mathcal{L}_l + \beta \cdot (\mathcal{L}_{sw12} + \mathcal{L}_{sw21}))$ & \textbf{95.16} & \textbf{94.24} & \textbf{97.50} \\
\hline
\end{tabularx}
\caption{Different LAC Loss Formulations on Model Performance}
\label{tab_lac_loss}
}
\end{table}

\noindent \textbf{Network Architecture.} Table \ref{tab_lac_arch} demonstrates that our performance evaluation across various network architectures highlights the superiority of the ResNet-50 model with pretrained weights combined with a Transformer.

\noindent \textbf{Hyperparameter of LAC Loss.} Table \ref{tab_lac_y} presents the optimal performance results obtained at the smoothing parameter $\gamma = 0.8$. 
Table~\ref{tab_lac_loss} demonstrates that adding the local alignment loss ($\mathcal{L}_l$) to the contrastive loss ($\mathcal{L}_c$), the classification accuracy increases by 1.72, progression accuracy by 2.12, and the Tau score by 2.57.

\section{Conclusion}
The paper introduces a novel approach to representation learning through a Local-Alignment Contrastive (LAC) loss that integrates a differentiable local alignment loss with a contrastive loss, all within a self-supervised framework. 
The method employs a differentiable affine Smith-Waterman algorithm to enable temporal alignment that dynamically adjusts to variations in action sequences. 
This approach is distinct in its focus on capturing local temporal dependencies and enhancing the discriminative learning of video embeddings, accommodating differences in action lengths and sequences.
Experimental results on the Pouring and PennAction datasets showcase the method's superior performance over existing state-of-the-art approaches.

\bibliographystyle{unsrt}  
\bibliography{references}  

\begin{thebibliography}{10}

\bibitem{2016_tsn}
Limin Wang, Yuanjun Xiong, Zhe Wang, Yu~Qiao, Dahua Lin, Xiaoou Tang, and Luc Van~Gool.
\newblock Temporal segment networks: Towards good practices for deep action recognition.
\newblock In {\em European conference on computer vision}, pages 20--36. Springer, 2016.

\bibitem{2017_quovadis}
Joao Carreira and Andrew Zisserman.
\newblock Quo vadis, action recognition? a new model and the kinetics dataset.
\newblock In {\em proceedings of the IEEE Conference on Computer Vision and Pattern Recognition}, pages 6299--6308, 2017.

\bibitem{2018_trn}
Bolei Zhou, Alex Andonian, Aude Oliva, and Antonio Torralba.
\newblock Temporal relational reasoning in videos.
\newblock In {\em Proceedings of the European conference on computer vision (ECCV)}, pages 803--818, 2018.

\bibitem{2019_slowfast}
Christoph Feichtenhofer, Haoqi Fan, Jitendra Malik, and Kaiming He.
\newblock Slowfast networks for video recognition.
\newblock In {\em Proceedings of the IEEE/CVF international conference on computer vision}, pages 6202--6211, 2019.

\bibitem{2019_actiontransnet}
Rohit Girdhar, Joao Carreira, Carl Doersch, and Andrew Zisserman.
\newblock Video action transformer network.
\newblock In {\em Proceedings of the IEEE/CVF conference on computer vision and pattern recognition}, pages 244--253, 2019.

\bibitem{2020_x3d}
Christoph Feichtenhofer.
\newblock X3d: Expanding architectures for efficient video recognition.
\newblock In {\em Proceedings of the IEEE/CVF conference on computer vision and pattern recognition}, pages 203--213, 2020.

\bibitem{2021_vit}
Alexey Dosovitskiy, Lucas Beyer, Alexander Kolesnikov, Dirk Weissenborn, Xiaohua Zhai, Thomas Unterthiner, Mostafa Dehghani, Matthias Minderer, Georg Heigold, Sylvain Gelly, et~al.
\newblock An image is worth 16x16 words: Transformers for image recognition at scale.
\newblock {\em arXiv preprint arXiv:2010.11929}, 2020.

\bibitem{2021_vivit}
Anurag Arnab, Mostafa Dehghani, Georg Heigold, Chen Sun, Mario Lu{\v{c}}i{\'c}, and Cordelia Schmid.
\newblock Vivit: A video vision transformer.
\newblock In {\em Proceedings of the IEEE/CVF international conference on computer vision}, pages 6836--6846, 2021.

\bibitem{2021_videoclass_actionclip}
Mengmeng Wang, Jiazheng Xing, and Yong Liu.
\newblock Actionclip: A new paradigm for video action recognition.
\newblock {\em arXiv preprint arXiv:2109.08472}, 2021.

\bibitem{2024_astar}
Myeongjun Kim, Federica Spinola, Philipp Benz, and Tae-Hoon Kim.
\newblock A*: Atrous spatial temporal action recognition for real time applications.
\newblock In {\em Proceedings of the IEEE/CVF Winter Conference on Applications of Computer Vision}, pages 7014--7024, 2024.

\bibitem{2017_tcn}
Pierre Sermanet, Corey Lynch, Yevgen Chebotar, Jasmine Hsu, Eric Jang, Stefan Schaal, Sergey Levine, and Google Brain.
\newblock Time-contrastive networks: Self-supervised learning from video.
\newblock In {\em 2018 IEEE international conference on robotics and automation (ICRA)}, pages 1134--1141. IEEE, 2018.

\bibitem{2019_tcc}
Debidatta Dwibedi, Yusuf Aytar, Jonathan Tompson, Pierre Sermanet, and Andrew Zisserman.
\newblock Temporal cycle-consistency learning.
\newblock In {\em Proceedings of the IEEE/CVF conference on computer vision and pattern recognition}, pages 1801--1810, 2019.

\bibitem{2021_lav}
Sanjay Haresh, Sateesh Kumar, Huseyin Coskun, Shahram~N Syed, Andrey Konin, Zeeshan Zia, and Quoc-Huy Tran.
\newblock Learning by aligning videos in time.
\newblock In {\em Proceedings of the IEEE/CVF Conference on Computer Vision and Pattern Recognition}, pages 5548--5558, 2021.

\bibitem{2021_gta}
Isma Hadji, Konstantinos~G Derpanis, and Allan~D Jepson.
\newblock Representation learning via global temporal alignment and cycle-consistency.
\newblock In {\em Proceedings of the IEEE/CVF Conference on Computer Vision and Pattern Recognition}, pages 11068--11077, 2021.

\bibitem{2022_carl}
Minghao Chen, Fangyun Wei, Chong Li, and Deng Cai.
\newblock Frame-wise action representations for long videos via sequence contrastive learning.
\newblock In {\em Proceedings of the IEEE/CVF Conference on Computer Vision and Pattern Recognition}, pages 13801--13810, 2022.

\bibitem{2024_lrprop}
Guy Bar-Shalom, George Leifman, and Michael Elad.
\newblock Weakly-supervised representation learning for video alignment and analysis.
\newblock In {\em Proceedings of the IEEE/CVF Winter Conference on Applications of Computer Vision}, pages 6909--6919, 2024.

\bibitem{1970_nw}
Saul~B Needleman and Christian~D Wunsch.
\newblock A general method applicable to the search for similarities in the amino acid sequence of two proteins.
\newblock {\em Journal of molecular biology}, 48(3):443--453, 1970.

\bibitem{2007_dtw}
Meinard M{\"u}ller.
\newblock Dynamic time warping.
\newblock {\em Information retrieval for music and motion}, pages 69--84, 2007.

\bibitem{2020_diff_nw}
James~T Morton, Charlie~EM Strauss, Robert Blackwell, Daniel Berenberg, Vladimir Gligorijevic, and Richard Bonneau.
\newblock Protein structural alignments from sequence.
\newblock {\em BioRxiv}, pages 2020--11, 2020.

\bibitem{2017_softdtw}
Marco Cuturi and Mathieu Blondel.
\newblock Soft-dtw: a differentiable loss function for time-series.
\newblock In {\em International conference on machine learning}, pages 894--903. PMLR, 2017.

\bibitem{1981_sw}
Temple~F Smith, Michael~S Waterman, et~al.
\newblock Identification of common molecular subsequences.
\newblock {\em Journal of molecular biology}, 147(1):195--197, 1981.

\bibitem{2013_pennaction}
Weiyu Zhang, Menglong Zhu, and Konstantinos~G Derpanis.
\newblock From actemes to action: A strongly-supervised representation for detailed action understanding.
\newblock In {\em Proceedings of the IEEE international conference on computer vision}, pages 2248--2255, 2013.

\bibitem{2014_vc_karpathy}
Andrej Karpathy, George Toderici, Sanketh Shetty, Thomas Leung, Rahul Sukthankar, and Li~Fei-Fei.
\newblock Large-scale video classification with convolutional neural networks.
\newblock In {\em Proceedings of the IEEE conference on Computer Vision and Pattern Recognition}, pages 1725--1732, 2014.

\bibitem{2014_vc_simonyan}
Karen Simonyan and Andrew Zisserman.
\newblock Two-stream convolutional networks for action recognition in videos.
\newblock {\em Advances in neural information processing systems}, 27, 2014.

\bibitem{2017_actiondetect}
Yue Zhao, Yuanjun Xiong, Limin Wang, Zhirong Wu, Xiaoou Tang, and Dahua Lin.
\newblock Temporal action detection with structured segment networks.
\newblock In {\em Proceedings of the IEEE international conference on computer vision}, pages 2914--2923, 2017.

\bibitem{2018_videocapsulenet}
Kevin Duarte, Yogesh Rawat, and Mubarak Shah.
\newblock Videocapsulenet: A simplified network for action detection.
\newblock {\em Advances in neural information processing systems}, 31, 2018.

\bibitem{2018_bsn}
Tianwei Lin, Xu~Zhao, Haisheng Su, Chongjing Wang, and Ming Yang.
\newblock Bsn: Boundary sensitive network for temporal action proposal generation.
\newblock In {\em Proceedings of the European conference on computer vision (ECCV)}, pages 3--19, 2018.

\bibitem{2022_actionformer}
Chen-Lin Zhang, Jianxin Wu, and Yin Li.
\newblock Actionformer: Localizing moments of actions with transformers.
\newblock In {\em European Conference on Computer Vision}, pages 492--510. Springer, 2022.

\bibitem{1994_opticalflow}
John~L Barron, David~J Fleet, and Steven~S Beauchemin.
\newblock Performance of optical flow techniques.
\newblock {\em International journal of computer vision}, 12:43--77, 1994.

\bibitem{1998_condensation}
Michael Isard and Andrew Blake.
\newblock Condensation—conditional density propagation for visual tracking.
\newblock {\em International journal of computer vision}, 29(1):5--28, 1998.

\bibitem{2005_stips}
Ivan Laptev.
\newblock On space-time interest points.
\newblock {\em International journal of computer vision}, 64:107--123, 2005.

\bibitem{2023_sw1}
Samantha Petti, Nicholas Bhattacharya, Roshan Rao, Justas Dauparas, Neil Thomas, Juannan Zhou, Alexander~M Rush, Peter Koo, and Sergey Ovchinnikov.
\newblock End-to-end learning of multiple sequence alignments with differentiable smith--waterman.
\newblock {\em Bioinformatics}, 39(1):btac724, 2023.

\bibitem{2023_sw2}
Felipe Llinares-L{\'o}pez, Quentin Berthet, Mathieu Blondel, Olivier Teboul, and Jean-Philippe Vert.
\newblock Deep embedding and alignment of protein sequences.
\newblock {\em Nature methods}, 20(1):104--111, 2023.

\bibitem{2020_simclr}
Ting Chen, Simon Kornblith, Mohammad Norouzi, and Geoffrey Hinton.
\newblock A simple framework for contrastive learning of visual representations.
\newblock In {\em International conference on machine learning}, pages 1597--1607. PMLR, 2020.

\bibitem{2015_resnet}
Kaiming He, Xiangyu Zhang, Shaoqing Ren, and Jian Sun.
\newblock Deep residual learning for image recognition.
\newblock In {\em Proceedings of the IEEE conference on computer vision and pattern recognition}, pages 770--778, 2016.

\end{thebibliography}

\clearpage
\newpage
\setcounter{section}{0}

\newcommand{\supplementarysection}[1]{%
    \refstepcounter{section} 
    \section*{\Alph{section} \quad #1}
    \addcontentsline{toc}{section}{Supplementary \Alph{section}: #1} 
}

\section*{Supplementary Materials}
\supplementarysection{Architecture Details}

The complete architecture of our encoder is shown in Table \ref{tab: lac_encoder}. 
The encoder consists of two main components: the \textit{Spatial Encoder} and the \textit{Temporal Encoder}.
The \textit{Spatial Encoder} is responsible for extracting spatial features from individual frames using ResNet-50 \cite{2015_resnet} with pretrained weights. 
This extraction process involves a series of convolutional layers, batch normalization, ReLU activations, max pooling, and bottleneck layers.
Subsequently, the \textit{Temporal Encoder} processes these spatial features across multiple frames to capture temporal dependencies. 
It employs fully connected layers with dropout and ReLU activations to enhance feature learning and prevent overfitting. 
A positional encoder is integrated to incorporate sequence information, and transformer encoder layers are utilized to model temporal relationships effectively. 
The final linear layer reduces the dimensionality of the embeddings to 128.

\begin{table}[h]
\centering
\small
\begin{tabularx}{\textwidth}{|l|l|X|c|}
\hline
\textbf{Model} & \textbf{Layer} & \textbf{Parameters} & \textbf{Output Size} \\ \hline

\multirow{8}{*}{Spatial Encoder} 
    & Conv2d & $ 7x7, 64$ & $112 \times 112 \times 64$ \\ \cline{2-4}
    & BatchNorm2d + ReLU & - & \\ \cline{2-4}
    & MaxPool2d & $3x3$ & $56 \times 56 \times 64$ \\ \cline{2-4}
    & Bottleneck conv2\_x & $\left[\begin{array}{c} 1 \times 1, 64 \\ 3 \times 3, 64 \\ 1 \times 1, 256 \end{array}\right] \times 3$ & $56 \times 56 \times 256$ \\ \cline{2-4}
    & Bottleneck conv3\_x & $\left[\begin{array}{c} 1 \times 1, 128 \\ 3 \times 3, 128 \\ 1 \times 1, 512 \end{array}\right] \times 4$ & $28 \times 28 \times 512$ \\ \cline{2-4}
    & Bottleneck conv4\_x & $\left[\begin{array}{c} 1 \times 1, 256 \\ 3 \times 3, 256 \\ 1 \times 1, 1024 \end{array}\right] \times 3$ & $14 \times 14 \times 1024$ \\ \cline{2-4}
    & Conv3d & $ 3x3x3, 512$ & $2 \times 10 \times 10 \times 512$ \\ \cline{2-4}
    & BatchNorm3d + ReLU & - & \\ \cline{2-4}
    & AdaptiveMaxPool3d & 1 & $1 \times 1 \times 1 \times 512$ \\ \hline

\multirow{6}{*}{Temporal Encoder} 
    & Dropout & $\text{p}=0.1$ & $64 \times 512$ \\ \cline{2-4}
    & Linear & 512 & \\ \cline{2-4}
    & ReLU & - & \\ \cline{2-4}
    & Linear & 256 & $64 \times 256$ \\ \cline{2-4}
    & Positional Encoder & - & $2 \times 32 \times 256$ \\ \cline{2-4}
    & Transformer Encoder & $\left[\begin{array}{c} 256, 8, 1024 \end{array}\right] \times 2$ & $2 \times 32 \times 256$ \\ \cline{2-4}
    & Linear & 128 & $64 \times 128$ \\ \hline

\end{tabularx}
\caption{Summarized Composition of the Spatial and Temporal Encoder Architectures. The parameter structure is as follows: For convolution layers, the format is kernel size and number of output channels (e.g., $7 \times 7, 64$). For pooling layers, the format is kernel size (e.g., $3 \times 3$). For other layers, specific parameters are listed (e.g., $p=0.1$ for dropout). For the Transformer Encoder, the format is the dimension of input and output vectors, number of attention heads, and dimension of the feedforward network hidden layer (e.g., $\left[\begin{array}{c} 256, 8, 1024 \end{array}\right] \times 2$). }
\label{tab: lac_encoder}
\end{table}

\supplementarysection{Differentiable Local-Alignment in Detail}

We provide a detailed derivation of the backward recursion used in differentiable local alignment algorithms. Following the approach described in \cite{2023_sw2}, we transform the discrete max function into a differentiable form, specifically a smooth max function. This smoothe max function is mathematically represented as follows:

\begin{equation}
  \max{}^{\gamma}(u_1, \ldots, u_n) := \gamma \, \log \, \sum_{i=1}^{n} \, e^{u_i/\gamma} ,
  \label{eq: smooth_max}
\end{equation}

Equation \ref{eq: smooth_max} enables differentiation that results in a softmax function, detailed as follows:

\begin{align}
    \frac{\partial \max{}^\gamma(\{u_1,\ldots, u_n\})}{\partial u_i} &= \exp\left(\frac{u_i - \max{}^\gamma(\{u_1,\ldots,u_n\})}{\gamma}\right) \\
    &= \exp\left(\frac{u_i - \gamma \log \left(\sum_{j=1}^n \exp\left(\frac{u_j}{\gamma}\right)\right)}{\gamma}\right) \\
    &= \exp\left(\frac{u_i}{\gamma} - \log \left(\sum_{j=1}^n \exp\left(\frac{u_j}{\gamma}\right)\right)\right) \\
    &= \exp\left(\frac{u_i}{\gamma}\right) \cdot \exp\left(-\log \left(\sum_{j=1}^n \exp\left(\frac{u_j}{\gamma}\right)\right)\right) \\
    &= \frac{\exp\left(\frac{u_i}{\gamma}\right)}{\sum_{j=1}^n \exp\left(\frac{u_j}{\gamma}\right)}
    \label{eq: partial_softmax}
\end{align}

This transformation makes the function differentiable, allowing the use of gradient-based optimization in alignment algorithms. The Smith-Waterman (SW) local alignment algorithm \cite{1981_sw} can be defined using dynamic programming as follows:

\begin{equation}
    D_{i,j} = S_{i, j} + \max \left\{
    \begin{array}{l}
        0 \\
        D_{i - 1,j - 1} \\
        I_{x_{i - 1,j - 1}} \\
        I_{y_{i - 1,j - 1}}
    \end{array} \label{eq: la_D}
\right. \\
\end{equation}

\begin{equation}
I_{x_{i,j}} = \max \left\{
    \begin{array}{l}
        D_{i, j - 1} - g_{o_{i, j}} \\
        I_{x_{i, j -1}} - g_{e_{i, j}}
    \end{array}
\right., \quad
I_{y_{i,j}} = \max \left\{
    \begin{array}{l}
        D_{i - 1, j} - g_{o_{i, j}}  \\
        I_{x_{i - 1, j}} - g_{o_{i, j}} \\
        I_{y_{i - 1, j}} - g_{e_{i, j}} 
    \end{array}
\right. \label{eq: la_Ix}
\end{equation}

where \(D_{i,j}\) represents the score of the optimal alignment ending at positions \(i\) and \(j\) in the two sequences being aligned. The scores for initiating or extending a gap are handled by \(I_{x_{i,j}}\) and \(I_{y_{i,j}}\). Additionally, \(g_{o_{i, j}}\) and \(g_{e_{i, j}}\) denote the penalties for opening and extending gaps. 
This method enables differentiation of the smoothed maximum score, denoted as $s* = \max^\gamma(0, \max^\gamma(D))$, with respect to \(S\), \(g_o\), and \(g_e\), treating these as parameters sensitive to derivatives.
The differentiation of the smoothed maximum score with respect to \(S\) can be defined as follows:

\begin{equation}
\begin{aligned}
    \frac{\partial s^*}{\partial S_{i,j}} 
    &= \left(\frac{\partial s^*}{\partial D_{i,j}}\right) \frac{\partial D_{i,j}}{\partial S_{i,j}}
\end{aligned}
\label{eq: la_bw_D}
\end{equation}

Using the inverse relationship defined in Eq. \ref{eq: la_Ix}, the differentiation of the smoothed maximum score with respect to \(g_o\) is described as follows:
\begin{align}
\frac{\partial s^*}{\partial g_{o_{i,j}}} 
    & =  \frac{\partial s^*}{\partial I_{x_{i,j}}} \, \frac{\partial I_{x_{i,j}}}{\partial g_{o_{i,j}}} 
    + \frac{\partial s^*}{\partial I_{y_{i,j}}} \, \frac{\partial I_{y_{i,j}}}{\partial g_{o_{i,j}}} \\
    & =  \frac{\partial s^*}{\partial I_{x_{i,j}}} \, \frac{\partial \max^\gamma \left(
        D_{i, j - 1} - g_{o_{i, j}},
        I_{x_{i, j -1}} - g_{e_{i, j}}
    \right)}{\partial g_{o_{i,j}}} \\
    & \quad + \frac{\partial s^*}{\partial I_{y_{i,j}}} \, \frac{ \partial \max^\gamma \left(
        D_{i - 1, j} - g_{o_{i, j}}, 
        I_{x_{i - 1, j}} - g_{o_{i, j}}, 
        I_{y_{i - 1, j}} - g_{e_{i, j}}  \right)
    }{\partial g_{o_{i,j}}} \\
    & = \frac{\partial s^*}{\partial I_{x_{i,j}}} \left( -\frac{\partial I_{x_{i,j}}}{\partial D_{i,j-1}} \right) 
    + \frac{\partial s^*}{\partial I_{y_{i,j}}} \left( -\frac{\partial I_{y_{i,j}}}{\partial D_{i-1,j}} - \frac{\partial I_{y_{i,j}}}{\partial I_{x_{i-1,j}}} \right)
 \label{eq: la_bw_go}
\end{align}

Similar to the above, using the inverse relationship from Eq. \ref{eq: la_Ix}, the differentiation of the smoothed maximum score with respect to \(g_e\) is defined as follows:

\begin{align}
\frac{\partial s^*}{\partial g_{e_{i,j}}} 
    & = \frac{\partial s^*}{\partial I_{x_{i,j}}} \frac{\partial I_{x_{i,j}}}{\partial g_{e_{i,j}}} 
    + \frac{\partial s^*}{\partial I_{y_{i,j}}} \frac{\partial I_{y_{i,j}}}{\partial g_{e_{i,j}}} \\
    & = \frac{\partial s^*}{\partial I_{x_{i,j}}} \frac{\partial \max^\gamma \left(
        D_{i, j - 1} - g_{o_{i, j}},
        I_{x_{i, j -1}} - g_{e_{i, j}}
    \right)}{\partial g_{e_{i,j}}} \\
    & \quad + \frac{\partial s^*}{\partial I_{y_{i,j}}} \frac{\partial \max^\gamma \left(
        D_{i - 1, j} - g_{o_{i, j}}, 
        I_{x_{i - 1, j}} - g_{o_{i, j}}, 
        I_{y_{i - 1, j}} - g_{e_{i, j}}  \right)
    }{\partial g_{e_{i,j}}} \\
    & = \frac{\partial s^*}{\partial I_{x_{i,j}}} \left( -\frac{\partial I_{x_{i,j}}}{\partial I_{x_{i,j-1}}} \right) 
    + \frac{\partial s^*}{\partial I_{y_{i,j}}} \left( - \frac{\partial I_{y_{i,j}}}{\partial I_{y_{i-1,j}}} \right)
 \label{eq: la_bw_ge}
\end{align}

Following \cite{2023_sw2}, we consider backtracking the SW algorithm to set the recursive calculation and see how changes in the current state propagate gradients forward to influence future states. We begin by deriving the expression for $\frac{\partial s^*}{\partial D_{i,j}} $ from Eq. \ref{eq: la_bw_D}, which can be obtained as follows:

\begin{align}
    \frac{\partial s^*}{\partial D_{i,j}} 
    &= \frac{\partial s^*}{\partial D_{i,j}} 
    + \frac{\partial s^*}{\partial D_{i+1,j+1}} \frac{\partial D_{i+1,j+1}}{\partial D_{i,j}}
    + \frac{\partial s^*}{\partial I_{x_{i,j+1}}} \frac{\partial I_{x_{i,j+1}}}{\partial D_{i,j}} 
    + \frac{\partial s^*}{\partial I_{y_{i+1,j}}} \frac{\partial I_{y_{i+1,j}}}{\partial D_{i,j}}
\end{align}

In this equation, \(\frac{\partial s^*}{\partial D_{i,j}}\) represents the immediate partial derivative of \(s^*\) with respect to \(D_{i,j}\), reflecting its direct impact. The term \(\frac{\partial s^*}{\partial D_{i+1,j+1}} \frac{\partial D_{i+1,j+1}}{\partial D_{i,j}}\) accounts for how \(D_{i,j}\) affects the future state \(D_{i+1,j+1}\). Similarly, \(\frac{\partial s^*}{\partial I_{x_{i,j+1}}} \frac{\partial I_{x_{i,j+1}}}{\partial D_{i,j}}\) reflects the influence on the future state \(I_{x_{i,j+1}}\), and \(\frac{\partial s^*}{\partial I_{y_{i+1,j}}} \frac{\partial I_{y_{i+1,j}}}{\partial D_{i,j}}\) reflects the influence on the future state \(I_{y_{i+1,j}}\).

From Equations \ref{eq: la_bw_go} and \ref{eq: la_bw_ge}, we see that the remaining partial derivatives to be calculated are $\frac{\partial s^*}{\partial I_{x_{i,j}}}$ and $\frac{\partial s^*}{\partial I_{y_{i,j}}}$. These derivatives can be derived recursively as follows:

\begin{align}
    \frac{\partial s^*}{\partial I_{y_{i,j}}}
    & = \frac{\partial s^*}{\partial D_{i+1,j+1} } \, \frac{\partial D_{i+1,j+1}}{\partial I_{y_{i,j}}}
    + \frac{\partial s^*}{\partial I_{y_{i+1,j}}}  \, \frac{\partial I_{y_{i+1,j}}}{\partial I_{y_{i,j}}}
\end{align}

\begin{align}
    \frac{\partial s^*}{\partial I_{x_{i,j}}}
    & = \frac{\partial s^*}{\partial D_{i+1,j+1} }  \, \frac{\partial D_{i+1,j+1}}{\partial I_{x_{i,j}}}
    + \frac{\partial s^*}{\partial I_{x_{i,j+1}}} \, \frac{\partial I_{x_{i,j+1}}}{\partial I_{x_{i,j}}} 
    + \frac{\partial s^*}{\partial I_{y_{i+1,j}}}\, \frac{\partial I_{y_{i+1,j}}}{\partial I_{x_{i,j}}}
\end{align}

where \(\frac{\partial s^*}{\partial I_{y_{i,j}}}\) accounts for the influence of \(I_{y_{i,j}}\) on the future states \(D_{i+1,j+1}\) and \(I_{y_{i+1,j}}\). Similarly, \(\frac{\partial s^*}{\partial I_{x_{i,j}}}\) reflects the influence of \(I_{x_{i,j}}\) on the future states \(D_{i,j+1}\) and \(I_{x_{i,j+1}}\). By recursively computing these partial derivatives, we can capture how changes in the current states \(D_{i,j}\), \(I_{x_{i,j}}\), and \(I_{y_{i,j}}\) propagate through the dynamic programming matrix.

\supplementarysection{More results}

In the main paper, the main result showcases the performance comparison of LAC and SOTA on the PennAction Dataset \cite{2013_pennaction}. The results represent the average across all 13 actions. Details of these actions can be found in Table \ref{tab: pn_action}.

\begin{table}[]
\small
\centering
\begin{tabularx}{\textwidth}{X|ccc|ccc|c|c}
\multirow{2}{*}{\textbf{Action}} & \multicolumn{3}{c|}{\textbf{Class}} & \multicolumn{3}{c|}{\textbf{AP@K}} & \multirow{2}{*}{\textbf{Progress}} & \multirow{2}{*}{\textbf{$\tau$}} \\ 
\cline{2-7}
 & \textbf{10} & \textbf{50} & \textbf{100} & \textbf{K=5} & \textbf{K=10} & \textbf{K=15} & & \\ 
\hline
\hline
baseball\_pitch & 92.50 & 90.95 & 90.68 & 91.48 & 91.41 & 90.95 &  93.28 & 88.64\\
baseball\_swing & 93.96 & 91.85 & 90.36 & 93.51 & 93.23 & 91.8 &  94.26 & 95.55\\
bench\_press & 94.83 & 91.64 & 91.17 & 93.76 & 92.88 & 91.53 &  96.64 & 93.42\\
bowl & 96.83 & 94.53 & 93.49 & 92.52 & 91.96 & 90.54 &  86.32 & 91.84\\
clean\_and\_jerk & 92.71 & 88.83 & 87.07  & 87.36 & 86.82 & 85.64 &  92.69 & 94.66\\
golf\_swing & 96.37 & 93.76 & 93.5 & 93.28 & 92.92 & 91.82 &  95.86 & 94.83\\
jumping\_jacks & 95.30 & 94.3 & 93.87 & 94.36 & 95.31 & 94.66 &  95.68 & 96.47\\
pushup & 94.04 & 94.84 & 94.38 & 94.66 & 94.38 & 94.23 &  96.8 & 96.64\\
pullup & 95.56 & 94.55 & 94.26 & 94.57 & 94.87 & 94.12 &  96.4 & 95.44\\
situp & 98.52 & 96.57 & 96.6 & 95.22 & 94.98 & 94.95 &  96.25 & 96.31\\
squat & 97.65 & 94.28 & 96.21 & 95.26 & 95.52 & 94.64 & 93.91 & 95.08\\
tennis\_forehand & 97.51 & 97.06 & 97.51 & 96.55 & 95.03 & 94.46 & 94.29 & 92.05\\
tennis\_serve & 96.62 & 96.09 & 95.09 & 97.76 & 95.01 & 95.10 & 92.33 & 92.35\\
\hline
\textbf{average} & \textbf{95.57} & \textbf{93.79} & \textbf{93.40} & \textbf{93.87} & \textbf{93.41} & \textbf{92.65} & \textbf{94.21} & \textbf{94.10} \\
\hline
\hline
\end{tabularx}
\caption{Performance comparison of LAC and SOTA on the PennAction Dataset \cite{2013_pennaction} across 13 actions.}
\label{tab: pn_action}
\end{table}






\end{document}